\documentclass[conference]{IEEEtran}
\IEEEoverridecommandlockouts
\usepackage[numbers,sort&compress]{natbib}
\usepackage{amsmath,amssymb,amsfonts}
\usepackage{algorithmic}
\usepackage{graphicx}
\usepackage{textcomp}
\usepackage{xcolor}
\usepackage{hyperref}
\hypersetup{
    colorlinks   = true,
    citecolor    = blue,
    urlcolor    =  blue
}
\usepackage{hyperref}
\usepackage{url}
\usepackage{balance}

\newcommand{\hf}[2]{\raisebox{-2.2pt}{\includegraphics[scale=0.09]{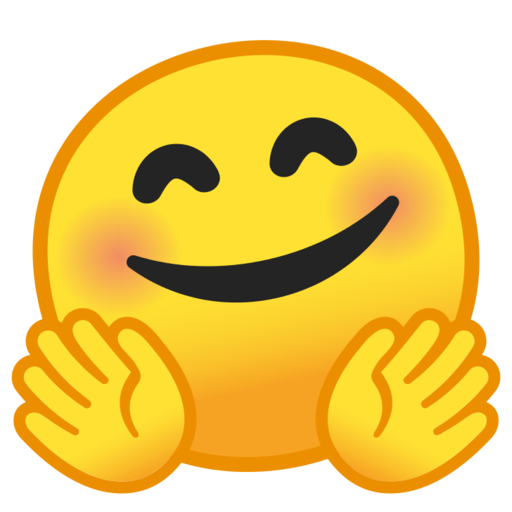}}~\href{#1}{\texttt{#2}}}

\newcommand{\gh}[2]{\raisebox{-2.2pt}{\includegraphics[scale=0.02]{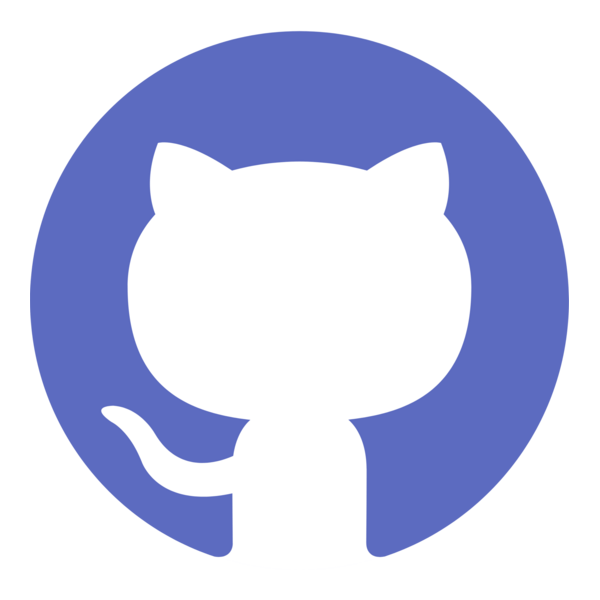}}~\href{#1}{\texttt{#2}}}

\def\BibTeX{{\rm B\kern-.05em{\sc i\kern-.025em b}\kern-.08em
    T\kern-.1667em\lower.7ex\hbox{E}\kern-.125emX}}
\begin{document}

\title{Golden Hour Divide: Trauma Care Accessibility and Resource Vulnerability in Sri Lanka}

\author{
\IEEEauthorblockN{%
Sonath Kirindage,
Vihanga Nimsara,
Sakindu Rajapaksa,
Kavyanga Hathurusinghe,
Lahiru Dilshan,\\
Subavarshana Arumugam,
Nathali Athukorala,
Sandareka Wickramanayake,
Nisansa de Silva}
\IEEEauthorblockA{Dept.\ of Computer Science \& Engineering, University of Moratuwa, Sri Lanka.\\
\texttt{\{sonathk.23, vihangan.23, sakindur.23, kavyangah.23, lahirud.23,}\\
\texttt{subavarshanaa.21, nathalia, sandarekaw, NisansaDds\}@cse.mrt.ac.lk}
}
}

\maketitle

\begin{abstract}
Timely intensive care dictates survival, yet emergency infrastructure remains unevenly distributed across Sri Lanka. While pre-hospital services have expanded, the transition to definitive care remains a critical bottleneck. This study evaluates national emergency resilience by quantifying the gap between clinical demand and the availability of specialized resources across all 25 districts. Using the latest national epidemiological data and terrain-aware H3 hexagonal modeling, we analyzed accessibility for seven critical conditions based on spatial gaps, clinical need-gaps, lethality, coverage, and resource availability. Based on these metrics, unsupervised K-Means clustering was applied to categorize districts into four policy-actionable archetypes: Critical Structural Exclusion, Institutional Mirages, Operational Capacity Strain, and High-Resilience Benchmarks. Our study suggests that severe service deficits exist in the Northern and Eastern provinces, where spatial gaps exceed 70\%, rendering the \textit{Golden Hour} operationally impossible. Notably, specialist scarcity drives systemic pressure more than bed capacity; underserved regions effectively function as institutional mirages. This study suggests that improving accessibility by 25\% in high-priority clusters would reduce the national need-gap by 9.65\%, providing a roadmap for the strategic redistribution of specialists to ensure healthcare equity.

\end{abstract}


\section{Introduction}

The \textit{Golden Hour} in emergency medicine refers to the critical 60-minute window during which timely clinical intervention has the greatest impact on survival~\cite{maurya2022neurotrauma}. It serves as a benchmark for evaluating the responsiveness and resilience of healthcare systems. In Sri Lanka, pre-hospital care has improved significantly with the introduction of the \textit{Suwa Seriya} island-wide ambulance service~\cite{thilakasiri20241990}. However, survival outcomes still depend on how quickly patients can transition from initial stabilization to definitive care.

Despite these advancements, disparities in access to critical care persist, particularly across regions with varying terrain and infrastructure~\cite{knowlton2017geospatial}. This study addresses these challenges through a data-driven framework that evaluates how effectively populations can reach life-saving medical services within the \textit{Golden Hour}, incorporating fine-grained spatial analysis using H3 hexagonal indexing system to better represent geographic variation~\cite{kmoch2022area}.

To capture these gaps, the study introduces three key measures. The Spatial Gap ($G_d$) represents the proportion of a district that falls outside the 60-minute access threshold, highlighting geographic limitations in timely care. The Need-Gap Index (NGI) reflects the imbalance between clinical demand and the availability of definitive-care resources, identifying areas where healthcare capacity is insufficient. The Lethality Ratio ($L_r$) measures deaths as a proportion of fatal outcomes, providing insight into the severity and effectiveness of care delivery.

By applying these concepts across major time-critical conditions, including cardiovascular and cerebrovascular emergencies, trauma, toxicological crises, and acute respiratory failures, this study offers a comprehensive view of definitive care accessibility. Positioned as a strategic planning tool, the framework highlights systemic inefficiencies and supports targeted improvements in national health infrastructure, with the goal of reducing preventable mortality and ensuring equitable access to care across all 25 districts of Sri Lanka~\cite{ouma2018access}.
\hf{https://huggingface.co/datasets/sonath0427/sri-lankan-medical-institutional-data}{Data} and \gh{https://github.com/sonath0427/golden-hour-divide}{code} for this work are publicly available.

\subsection{Research Hypotheses}

To validate this framework, the following hypotheses are tested:

\subsection*{Hypothesis 1: The Spatial Fragmentation Hypothesis}

\textbf{H0:} There is no significant direct proportional relationship between the reduction in spatial gap \textbf{($G_d$)} and the density of ICU facilities.

\textbf{H1:} The relationship between ICU density and spatial gap \textbf{($G_d$)} is statistically significant but weak, with terrain-induced constraints being the dominant driver of inaccessibility.

\subsection*{Hypothesis 2: The Infrastructure Mirage Hypothesis}

\textbf{H0:} Clinical outcomes ($L_r$) are not significantly more strongly associated with access to definitive care facilities (ICUs/OTs) than with the density of primary stabilization units (ETUs).

\textbf{H1:} Clinical outcomes ($L_r$) are significantly more strongly associated with access to definitive care facilities (ICUs/OTs) than with the density of primary stabilization units (ETUs).

\subsection*{Hypothesis 3: The Specialist Scarcity Hypothesis}

\textbf{H0:} Specialist availability is not a significantly stronger driver of systemic pressure (NGI) than physical bed capacity.

\textbf{H1:} Specialist availability is a significantly stronger driver of systemic pressure (NGI) than physical bed capacity.

\subsection*{Hypothesis 4: The Pareto Optimization Hypothesis}

\textbf{H0:} There is no significant difference in the reduction of the National Need-Gap Index between uniform resource distribution and targeted intervention in high-priority districts.

\textbf{H1:} Targeted intervention in high-priority districts produces a significantly greater reduction in the National Need-Gap Index than uniform resource distribution.

\section{Related Work and Literature Review
}
This section assesses emergency access through treatment times and mapping, reviewing the \textit{Golden Hour}, Sri Lankan trauma care, and geospatial health studies.

\subsection{The \textit{Golden Hour} and Definitive Care Paradigms}

The \textit{Golden Hour} concept suggests that survival is highest when definitive care is delivered within the first 60 minutes after the onset of injury or illness~\cite{babu2023golden}. In conditions such as neurotrauma, acute cardiovascular events, and severe envenomation, stabilization alone is insufficient without surgical intervention or intensive care~\cite{meara2015global}. Therefore, the effectiveness of the \textit{Golden Hour} depends less on distance to any facility and more on access to hospitals equipped with Operating Theatres (OTs) and Intensive Care Units (ICUs).

\subsection{ Sri Lankan Trauma System}

Sri Lanka has improved pre-hospital response through the Suwa Seriya service and the expansion of Emergency Treatment Units (ETUs) at Base Hospitals~\cite{o2017delivering}. However, trauma pathways remain fragmented, with no standardized national policy or integrated communication system~\cite{varathan2025advancing}. The concentration of specialists in urban centers further leads to referral delays for rural patients.
Audits~\cite{prarthana2025geospatial} report moderate-to-high inequality ($Gini = 0.41$) in ICU bed distribution, with 60\% of capacity concentrated in Colombo, Gampaha, and Kandy, which contain only 39\% of the population. Problems in emergency care come from poor resource distribution, not a lack of buildings.

\subsection{Geospatial Modeling and Temporal Latency}\label{AA}

Geographic Information Systems (GIS) are widely used to assess healthcare access. The Two-Step Floating Catchment Area (2SFCA) method~\cite{abeyrathna2025evaluating} shows that primary care in Anuradhapura is often adequate but depends on the private sector and is linked to socio-economic factors.
However, primary care proximity (5-10 km) does not reflect emergency access, where road conditions and facility capacity are critical. Recent data from 2026 show ambulance response times exceeding 35-45 minutes in rural and tourism-heavy regions in the East and North-Central provinces~\cite{Weerakoon2026Emergency}, while findings from The Asian Development Bank\footnote{\url{https://www.adb.org/projects/51107-002/main}} indicates doubled response times in mountainous districts such as Badulla. These delays significantly reduce the effective \textit{Golden Hour} before patients reach appropriate care.

\subsection{Synthesis and Research Gap}
The literature shows that timely intervention is the strongest predictor of survival~\cite{babu2023golden}. In Sri Lanka however, centralized resources~\cite{prarthana2025geospatial}, fragmented pathways~\cite{varathan2025advancing}, and pre-hospital delays in peripheral regions~\cite{Weerakoon2026Emergency} further limit access to definitive care.
This study addresses a key gap by focusing on definitive emergency care rather than primary care. Unlike prior work using Euclidean distance~\cite{sangasumana2019gis} or node-based connectivity analysis~\cite{mathivathany2015accessibility}, it applies the Uber H3 Hexagonal Index~\cite{kmoch2022area} for a road-network-based analysis. Traditional studies rely on square grids or administrative boundaries, which suffer from the Modifiable Areal Unit Problem (MAUP)~\cite{buzzelli2020modifiable} and cause directional bias. In contrast, the H3 hexagonal system minimizes these errors, ensures highly accurate neighbor spacing when modeling travel friction, and allows for much faster data aggregation than old raster models. It is the first study in Sri Lanka to combine seven time-critical conditions with high-resolution isochrones to derive a composite Need-Gap Index (NGI).

\section{ Dataset and Preprocessing}
To facilitate a multidimensional analysis of healthcare accessibility, this study acquired data from two primary sources: the Annual Health Bulletin (2024)\footnote{\url{https://www.health.gov.lk/wp-content/uploads/2026/03/Annual-Health-Bulletin-2024-compressed.pdf}} and the Medical Statistics Unit (MSU) of the Ministry of Health, Sri Lanka. These data reflect the actual hospital infrastructure and patient counts for the year 2024. The resulting dataset is categorized into four analytical pillars:

\subsection*{I. Demographic and Administrative Baseline}
The spatial framework is based on district-level demographic data, including total land area (km$^2$) and population density (persons per km$^2$). These metrics provide the denominator for population-normalization.

\subsection*{II. Clinical Demand and Epidemiological Profile}

Clinical demand was estimated using district-level morbidity and mortality records for seven time-critical conditions:

\begin{itemize}
\item Cardiovascular \& Cerebrovascular: Ischaemic heart disease and Cerebrovascular disease.
\item Respiratory: Asthma and Pneumonia.
\item Toxicological \& Environmental: Snake bites and acute poisoning (categorized by Organophosphates, Carbamates, and non-medicinal substances).
\item Traumatic: Road traffic accidents and other external traumatic injuries.
\end{itemize}

For each condition, Total Cases (Live Discharges + Deaths) were utilized as the primary incidence metric to measure systemic demand.

\subsection*{III. Institutional Supply and Specialist Resource Mapping}
The study analyzed tertiary and secondary institutions across all 25 districts, focusing on Base Hospitals and above within a national network of over 1,250 facilities. Capacity was measured via functional ICU bed and operating theatre counts. Key specialists, including general surgeons, neurosurgeons, cardiologists, physicians, and radiologists, were mapped to assess advanced clinical care availability.

\subsection*{IV. Geospatial Repository}
A spatial dataset of 83 healthcare facilities was developed using Google Maps Platform GPS coordinates. This supported OSRM-based isochrone analysis to estimate travel times and identify spatial accessibility gaps.
\subsection*{V. Data Preprocessing and Inclusion Criteria}
Null entries were treated as zero to reflect resource absence. The study focuses exclusively on public civilian secondary and tertiary hospitals; Apeksha Hospital and military/police facilities were excluded as they do not serve the general public emergency referral system.

\section{Methodology}

This study uses a four-phase framework to analyze ICU access and need across Sri Lanka’s 25 districts, combining terrain-aware mapping with disease demand and resource density.

\subsection{Phase 1: Terrain-Aware Spatial Gap ($G_d$)}
Geographic accessibility is modeled using an elevation-informed travel-time surface\footnote{\url{https://portal.opentopography.org/datasetMetadata?otCollectionID=OT.032021.4326.1}}. The foundational road network geometry and attribute data for Sri Lanka were retrieved from OpenStreetMap extracts\footnote{\url{https://download.geofabrik.de/asia/sri-lanka.html}}.

\textbf{Spatial Discretisation:} Sri Lanka is partitioned into Uber H3 hexagons (Resolution 8, $\sim$0.737 km$^2$) to capture fine-grained spatial variability.

\textbf{Terrain-Aware Speed Model:} Road segment slope is computed as:
\begin{equation}
\text{Grade} = \frac{|\text{End\_Elevation} - \text{Start\_Elevation}|}{\text{Segment\_Length}}
\end{equation}

Adjusted speed is derived from base speed ($V_{\text{base}}$) with a capped terrain penalty:
\begin{equation}
V_{\text{adj}} = \max\left( V_{\text{base}} \times \left(1 - \min(0.5 \times \text{grade\_percentage},0.75)\right),\ 5 \right)
\end{equation}

\textbf{Travel Time Estimation:} Travel time from each hexagon centroid to the nearest of the 83 ICU hospitals is computed using haversine distance ($D$) with a 1.35 detour factor~\cite{boscoe2012nationwide}:
\begin{equation}
T = \frac{D \times 1.35}{V_{\text{H3}}} \times 60
\end{equation}

Custom travel time methodologies utilizing open-source infrastructure provide critical advantages in financial scalability and methodological transparency, facilitating scientific reproducibility and the processing of large-scale matrices that are often prohibitively expensive or restricted by the quotas of commercial APIs.

\textbf{Spatial Interpolation:} A K-Nearest Neighbors (KNN) regressor is applied to interpolate data gaps in roadless regions (e.g., forests). This transformation converts discrete road-network samples into a continuous accessibility surface.

\textbf{Spatial Gap ($G_d$):} Proportion of district area exceeding the 60-minute \textit{Golden Hour}:
\begin{equation}
G_d = \frac{\text{Hexagons} > 60 \text{ min}}{\text{Total Hexagons in District}}
\end{equation}

\subsection{Phase 2: Infrastructure Paradox (Supply vs. Access)}
This phase evaluates the mismatch between stabilization capacity (ETUs) and definitive care (ICUs).

\textbf{Institutional Access Ratio (IAR):}
\begin{equation}
\text{IAR} = \frac{\text{ICU-Equipped Hospitals}}{\text{ETU-Capable Hospitals}}
\end{equation}

\subsection{Phase 3: Clinical Need-Gap Index (NGI) and Lethality Ratio ($L_r$)}
This phase integrates clinical demand with spatial and structural constraints.

\textbf{Lethality Ratio ($L_r$):} Represents the proportion of fatal outcomes, used to validate the clinical impact of accessibility gaps.
\begin{equation}
L_r = \frac{\text{Total Deaths}}{\text{Total Cases}}
\end{equation}

\textbf{Integrated Definitive Care Capacity (Resource Score):} The Resource Score weights are based on Resource Intensity. We assigned higher weights to specialized human capital (Specialists) and advanced life-support equipment (ICU/OTs) because these are the primary drivers of survival in critical cases, whereas basic infrastructure acts as a secondary support.
\begin{equation}
\text{Resource Score} = (w_1 \times \text{ICU\_Beds}) + (w_2 \times \text{OTs}) + (w_3 \times \text{Specialists})
\end{equation}
with weights: $w_1 = 0.2$, $w_2 = 0.3$, $w_3 = 0.5$.

\textbf{General NGI:} Measures systemic pressure by relating population demand, spatial barriers, and available resources.
\begin{equation}
\text{NGI} = \frac{(\text{Total Cases} / \text{Population}) \times G_d}{\text{Resource Score} / \text{Population}}
\end{equation}

Because the baseline hospital infrastructure and specialist counts are fixed for the year, the NGI uses total annual case volumes. This approach measures long-term, district-wide systemic pressure rather than daily operational shifts, providing a macro-level indicator of year-round resource deficits.

\subsubsection{Disease-Specific NGI}
To reflect clinical heterogeneity,   NGI is computed for seven time-critical conditions with adjusted resource components.

All components of the Adjusted Resource Score are Min-Max normalized (0-1) before weighting for comparability.

\textbf{Normalization:}
\begin{equation}
\text{Normalized Resource} = \frac{\text{Value} - \text{Min}}{\text{Max} - \text{Min}}
\end{equation}

\textbf{Disease-Specific NGI:} Quantifies condition-specific system strain by incorporating tailored resource requirements.
\begin{equation}
\text{NGI}_{\text{disease}} = \frac{(\text{Condition\_Cases} / \text{Population}) \times G_d}{\text{Adjusted\_Resource\_Score} / \text{Population}}
\end{equation}

The Adjusted Resource Score uses a Weighted Resource Intensity approach, merging AHP logic~\cite{saaty1987analytic} with the RIW methodology~\cite{knox2021patient}. As shown in Table~\ref{tab:resource_score_config} Assets are prioritized by their role in the Chain of Survival: Active Assets (Specialists) carry the highest weight (0.35-0.60), followed by Definitive Care (ICUs/OTs) and Primary Entry (ETUs). This shifts the focus from basic infrastructure counts to functional life-saving capability.

\begin{table*}[t]
\caption{Condition-Specific Resource Score Configuration}
\centering
\resizebox{\linewidth}{!}{
\begin{tabular}{|l|l|l|}
\hline
\textbf{Condition Category} & \textbf{Resource Score Components} & \textbf{Logic} \\
\hline

Ischaemic Heart Disease 
& Cardiologists(0.45) + Radiologists(0.15) + ETU(0.1) + ICU(0.2) + OT(0.1) 
& Requires both medical and potential surgical intervention. \\

Cerebrovascular Disease 
& Neurologists/Neurosurgeons(0.35) + Radiologists(0.25) + ETU(0.1) + ICU(0.2) + OT(0.1) 
& Critical for hemorrhagic stroke management. \\

Traumatic Injuries 
& Surgeons(Gen/Ortho/Neuro)(0.35) + Radiologists(0.15) + ETU(0.1) + ICU(0.2) + OT(0.2) 
& Primary requirement for definitive surgical care. \\

Snake Bites 
& General Physicians(0.5) + ETU(0.3) + ICU(0.2) 
& Medical management; OTs are not primary requirements. \\

Poisoning 
& Anaesthesiologists/General Physicians(0.6) + ETU(0.2) + ICU(0.2) 
& Focus on ventilation and toxicological stabilization. \\

Asthma 
& Chest Physicians(0.5) + Paediatricians(0.2) + ETU(0.2) + ICU(0.1) 
& Respiratory support is the primary limiting factor. \\

Pneumonia 
& Physicians(Chest/General)(0.5) + Radiologists(0.2) + ETU(0.2) + ICU(0.1) 
& Critical care monitoring and oxygenation focus. \\
\hline

\end{tabular}
}
\label{tab:resource_score_config}
\end{table*}

\subsection{Phase 4: Policy Clustering and System Optimisation}

Districts are clustered via K-Means using the feature vector 
$[G_d,\ \text{NGI},\ L_r,\ \text{TCR},\ \text{Resource Score}]$. The choice of $K=4$ was based on cluster stability, domain interpretability, and the need to maintain a high observation-to-parameter ratio given the small 25-district dataset. Evaluation of values from $K=3$ to $K=6$ showed that $K=3$ caused under-fitting and masked regional disparities, while $K \ge 5$ caused over-segmentation and unstable singleton clusters. Thus, $K=4$ was selected as the optimal threshold, providing mathematically stable groupings that capture four distinct health-system archetypes in Sri Lanka.

\textbf{Territorial Coverage Ratio (TCR):}

\begin{equation}
\text{TCR} = \left( \frac{\text{Covered Hexagon Area}}{\text{Total District Area}} \right) \times 100
\end{equation}

This metric represents geographic equity by measuring land coverage within the \textit{Golden Hour}.

\textbf{Cluster Archetypes:}

\begin{itemize}
\item \textbf{Cluster 1: Critical Structural Exclusion (Red)}  
High $G_d$ ($>50\%$), low TCR, and high $L_r$.

\item \textbf{Cluster 2: Institutional Mirages (Orange)}  
Low $G_d$ but high $L_r$ and low specialist density.

\item \textbf{Cluster 3: Operational Capacity Strain (Yellow)}  
Low $G_d$ but high NGI.

\item \textbf{Cluster 4: High-Resilience Benchmarks (Green)}  
Low $G_d$, high TCR, and low NGI. Represents optimized system performance.
\end{itemize}

\textbf{Optimization Simulation}: A statistical simulation was conducted to test high-priority interventions. By mathematically reducing the Spatial Gap ($G_d$) in Red Clusters by a 25\% factor, the model calculates the resulting decrease in the National NGI. This measures the systemic ``Return on Investment'' for expanding infrastructure within the most excluded regions.

\section{Experiments And Results}

To provide a clear overview of the data used in this study, Table~\ref{tab:stat_summary} presents the summary statistics for the 25 districts of Sri Lanka.
\begin{table}[htbp]
\caption{Statistical Summary of Healthcare Accessibility Features}
\centering 
\footnotesize 
\setlength{\tabcolsep}{3pt} 
\begin{tabular}{|l|c|c|c|c|}
\hline
\textbf{Feature} & \textbf{Mean} & \textbf{Standard Deviation} & \textbf{Maximum} & \textbf{Minimum} \\
\hline
$G_d$ & 44.9 & 29.0 & 84.1 & 0.2 \\
$L_r$ & 0.013 & 0.005 & 0.021 & 0.003 \\
Resource Score & 39.2 & 45.7 & 242.2 & 5.2 \\
Total Cases & 61784 & 42471.9 & 179759 & 6570 \\
\hline
\end{tabular}
\\[10pt]
\label{tab:stat_summary}
\end{table}

\subsection{Phase 1: Spatial Gap Analysis ($G_d$)}

\begin{figure}[htbp]
    \centering
    \includegraphics[width=\linewidth]{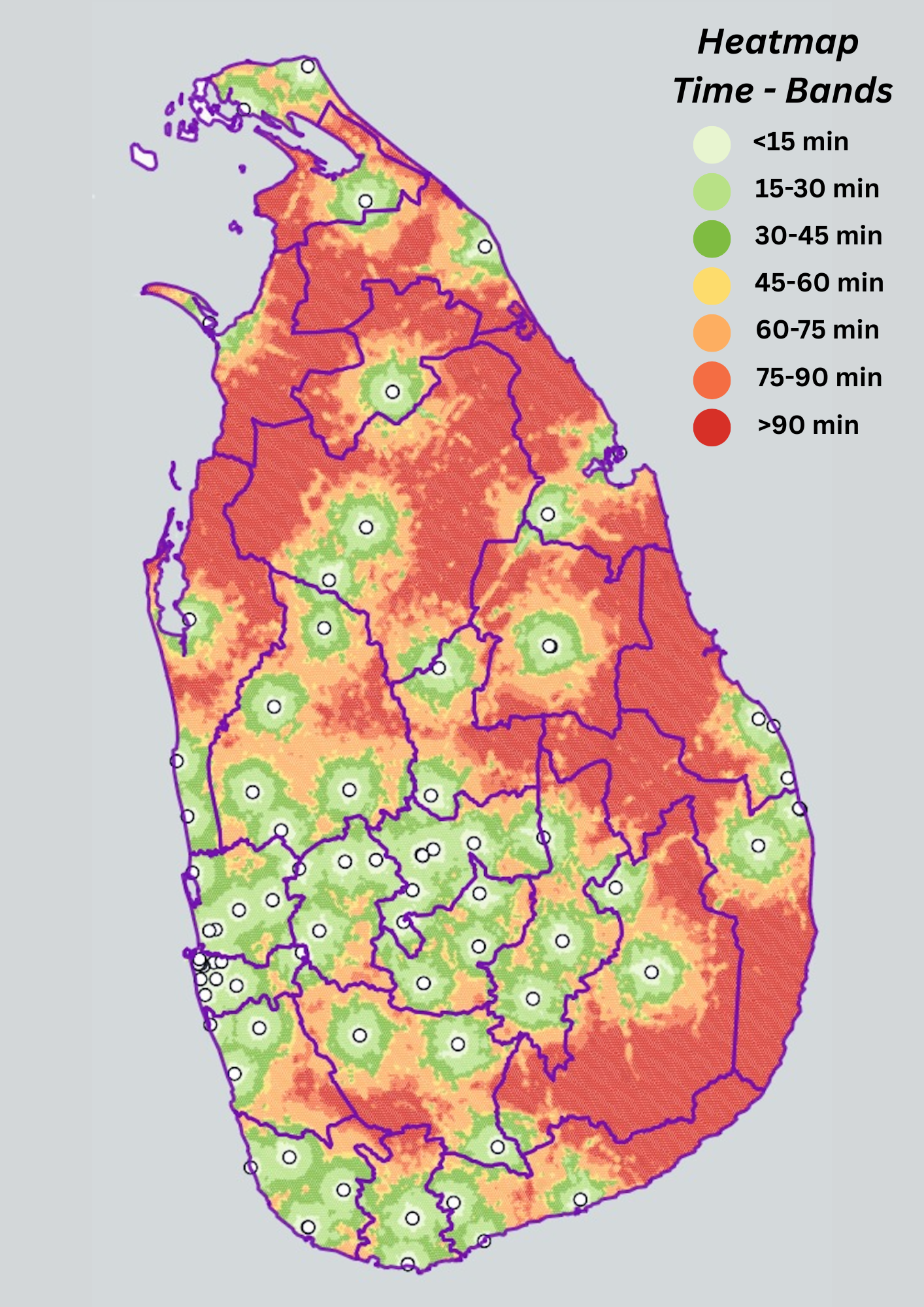}
    \caption{Terrain-Aware Spatial Gap ($G_d$) distribution across districts}
    \label{fig:gd_heatmap}
\end{figure}

The terrain-aware model (Fig.~\ref{fig:gd_heatmap}) reveals a clear geographic divide in line with the observations of previous mobility studies~\cite{lokanathan2016potential,lokanathan2014using} in Sri Lanka. In districts such as Mullaitivu ($75.30\%$) and Moneragala ($73.87\%$), $G_d$ exceeds $70\%$, making the \textit{Golden Hour} operationally impossible. In contrast, Colombo ($G_d = 0.23\%$) and Gampaha ($G_d = 0.74\%$) exhibit near-total coverage, while moderate gaps are observed in Galle ($13.33\%$) and Matara ($21.35\%$). Pearson correlation between hospital density and $G_d$ yielded $r = -0.53$ ($p = 0.006$), rejecting Hypothesis 1 (H0). The $R^2$ of 0.284 shows that hospital density explains only 28\% of the variance in $G_d$. This indicates that 72\% of the spatial gap is driven by structural factors, likely terrain and placement inefficiency, suggesting that infrastructure density alone cannot guarantee accessibility.

\subsection{Phase 2: The Infrastructure Mirage}

\begin{figure*}[htbp]
    \centering
    \includegraphics[width=0.8\linewidth]{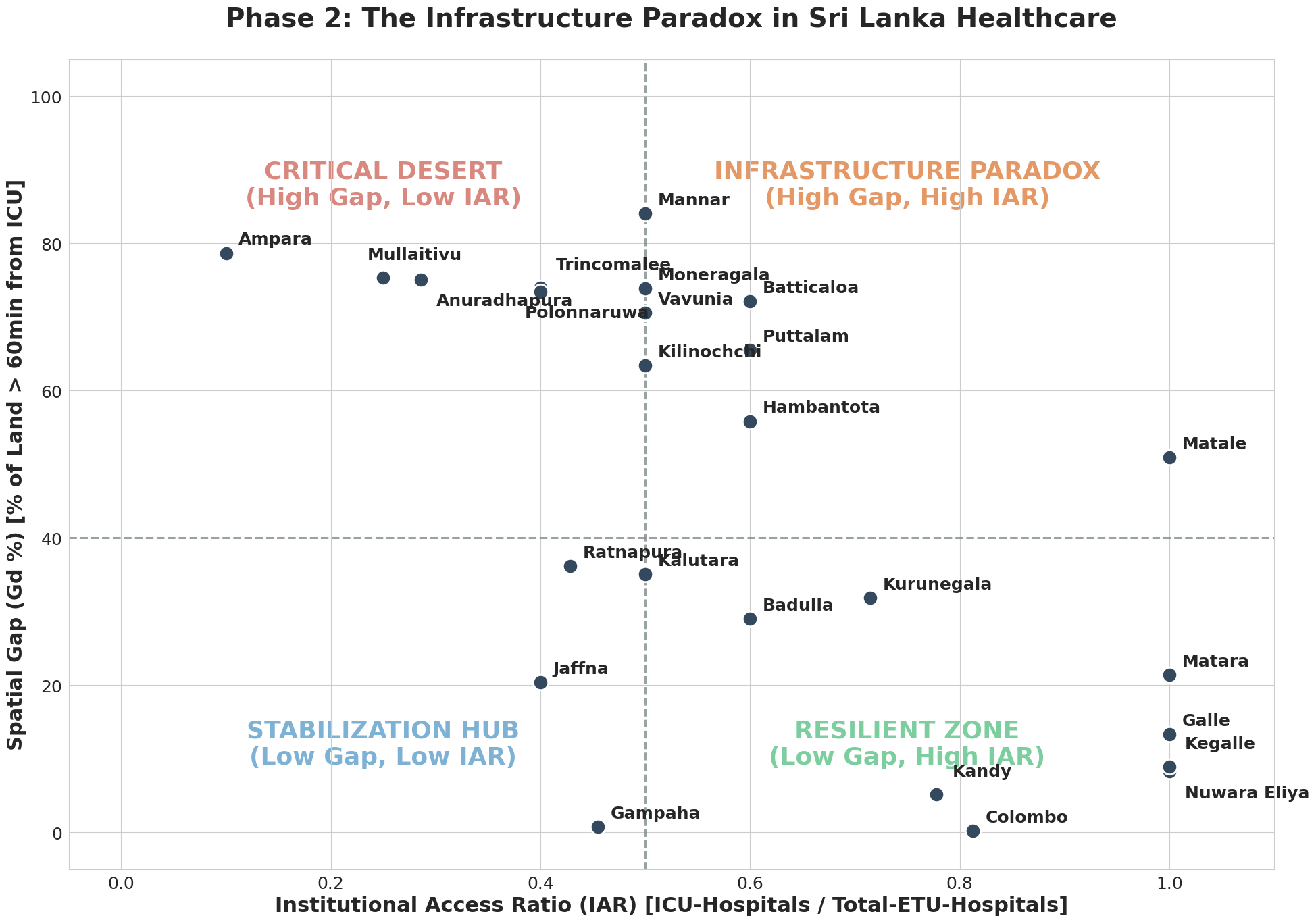}
    \caption{The Infrastructure Paradox in Sri Lanka Healthcare}
    \label{fig:Clinical Need-Gap}
\end{figure*}

Phase 2 identifies a deceptive sense of security in high-density primary facility zones lacking definitive care.
\subsubsection{Quadrant Analysis}
Districts are categorized into four archetypes (Fig.~\ref{fig:Clinical Need-Gap}): Critical Desert (Ampara, Mullaitivu), Infrastructure Paradox (Matale, Hambantota), Stabilization Hub/Mirage (Gampaha, Jaffna), and Resilient Zone (Colombo, Kandy).
\subsubsection{Validation of H2}
Correlation suggests definitive care access ($TCR$) is the engine of survival, rejecting Hypothesis 2 (H0). ETU Density vs. $L_r$ yielded $r = +0.4598$ ($p = 0.0208$), while $TCR$ vs. $L_r$ yielded a stronger $r = +0.7142$ ($p < 0.001$).

\subsection{Human Capital vs. Physical Infrastructure}
Phase~3 addresses the ``Specialist Scarcity'' bottleneck where human capital decouples from physical capacity.

\begin{figure*}[htbp]
    \centering
    \includegraphics[width=1\linewidth]{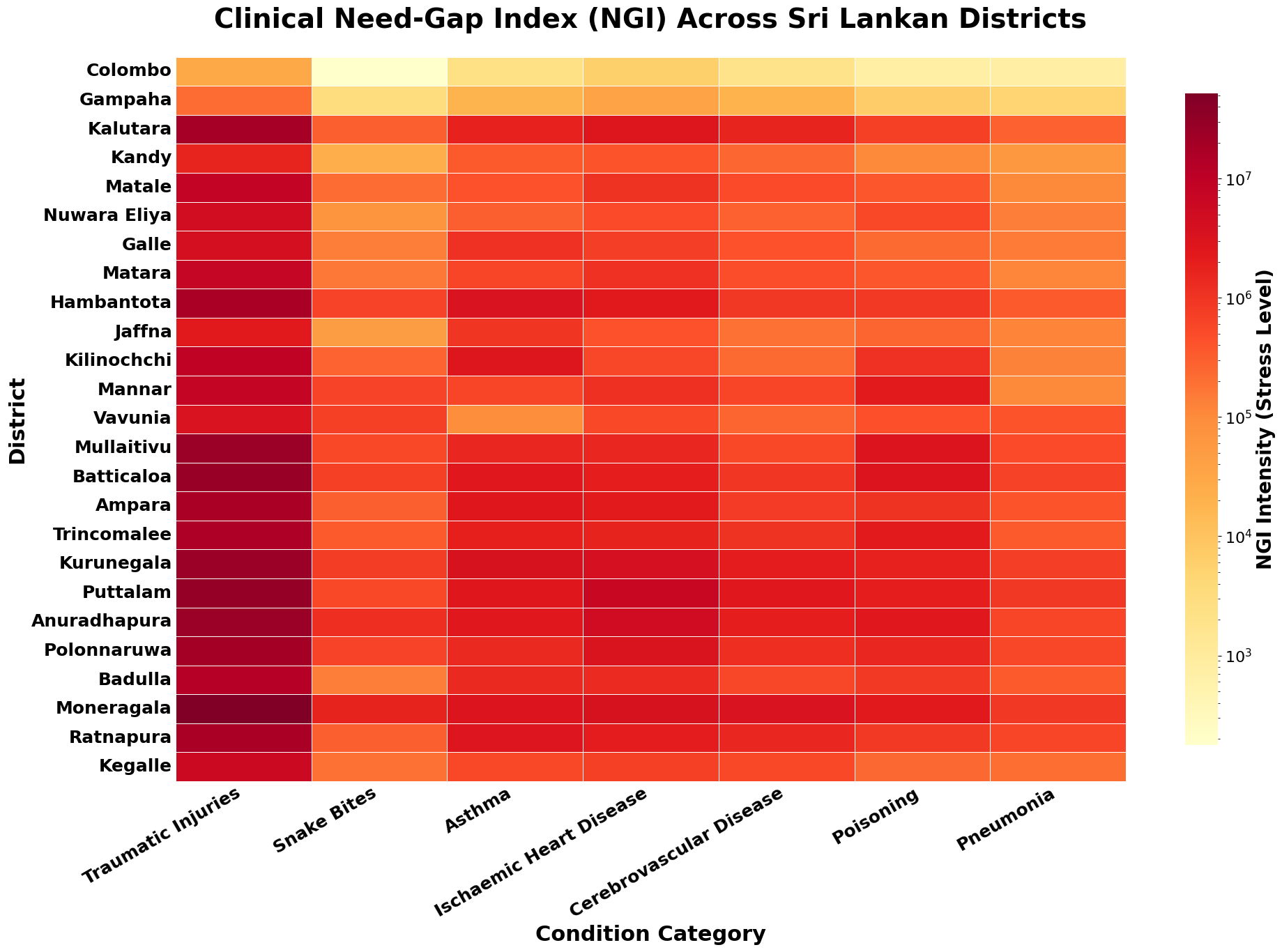}
    \caption{Clinical Need-Gap Index by Condition and District}
    \label{fig:placeholder}
\end{figure*}

\begin{enumerate}
    \item \textbf{General NGI Disparities:} Colombo’s NGI is $170.23$, whereas Moneragala ($215{,}292.71$) and Kurunegala ($198{,}481.63$) face pressure levels $1{,}000\times$ higher. This is driven by high numerators ($G_d$ and Cases) and low denominators (Resource Scores). For example, Mannar’s $G_d$ ($84.07\%$) and low Resource Score ($17.1$) vs. Colombo’s $G_d$ ($0.23\%$) and Resource Score ($242.2$) create these massive scale variances.
    
    \item \textbf{Validation of $H_3$:} Regression analysis confirmed Specialist Density is significant ($p = 0.047$), while Bed Density is not ($p = 0.214$). Spearman correlation suggests that a higher Specialist-to-Bed Ratio reduces $L_r$ ($r = -0.45$, $p = 0.023$).
    
    \item \textbf{The Specialist Bottleneck:} Localized vulnerabilities are extreme [Fig.~\ref{fig:placeholder}]. Moneragala peaks at a Trauma NGI of $52{,}003{,}876$, while categorical scarcity drives high pressure in Puttalam for IHD ($6.6\text{M}$) and Kurunegala for Asthma ($3.6\text{M}$).
\end{enumerate}

\subsection{Policy Archetypes: The Four-Quadrant Strategy}

\begin{figure*}[!htbp]
    \centering
    \includegraphics[width= 1\linewidth]{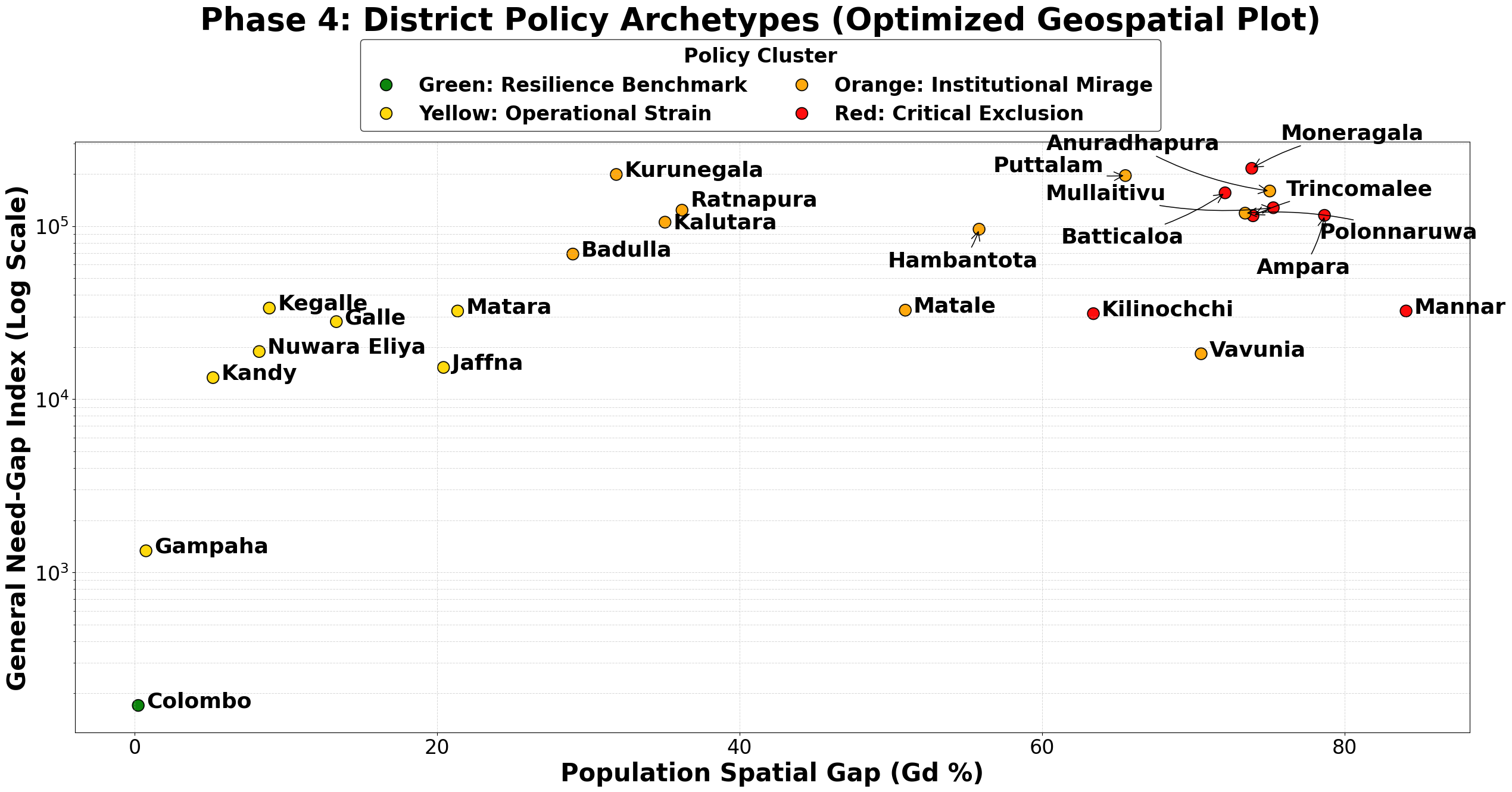}
    \caption{District Policy Archetypes}
    \label{fig:District Policy Archetypes}
\end{figure*}

Clustering Analysis Fig.~\ref{fig:District Policy Archetypes} provides a strategic roadmap for optimization:

\begin{itemize}
\item \textbf{Critical Exclusion (Red):} (Moneragala, Ampara) Dual exclusion of isolation ($G_d \approx 74.5\%$) and systemic inadequacy (NGI $> 113k$)
\item \textbf{Institutional Mirage (Orange):} (Kurunegala, Anuradhapura) Moderate proximity but low ``Active Assets'' (High NGI)
\item \textbf{Operational Strain (Yellow):} (Gampaha, Kandy) High accessibility ($G_d \approx 11.2\%$) but overwhelmed by volume
\item \textbf{Resilience Benchmark (Green):} (Colombo) Optimized transport and immediate definitive care
\end{itemize}

\subsection{National Impact and the Pareto Optimization Simulation}

\begin{figure}[!htbp]
    \centering
    \includegraphics[width=0.75\linewidth]{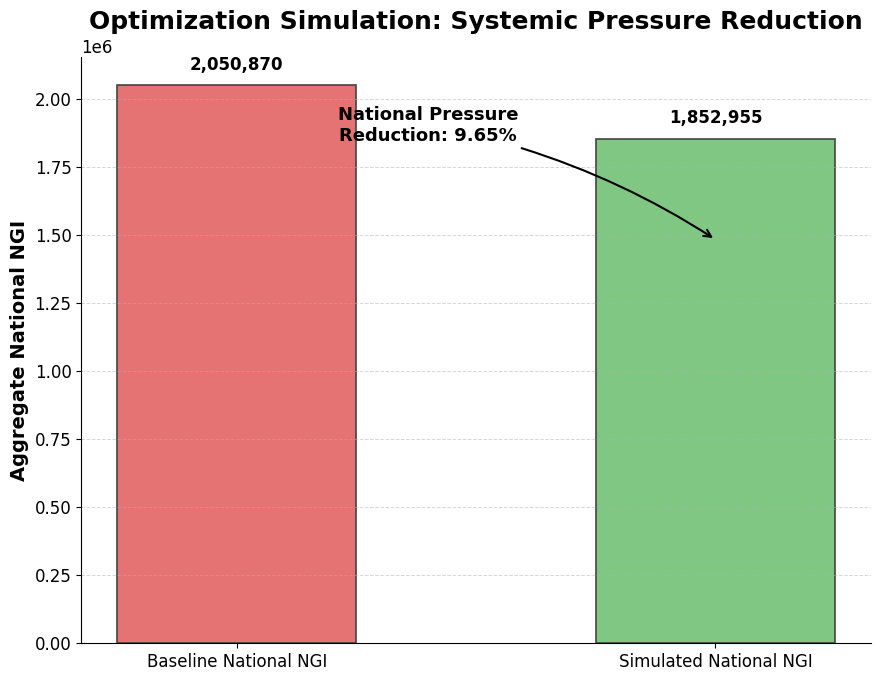}
    \caption{Simulated Pressure Reduction}
    \label{fig:Optimization Simulation}
\end{figure}

An Optimization Simulation Fig.~\ref{fig:Optimization Simulation} serves as prescriptive proof for H4 (Targeted Investment).

A 25\% increase in definitive care within 7 ``Red-Cluster'' districts reduced the National NGI by 9.65\%. This strongly suggests non-linear optimization: targeting the most excluded 20\% of land area yields a ~10\% national efficiency gain.

\section{Critical Discussion}

\begin{enumerate}

\item \textbf{The Referral Paradox and General NGI Dynamics}\\
Positive correlations between ETU density ($r = +0.4598$) and TCR ($r = +0.7142$) with the lethality ratio ($L_r$) reveal a referral paradox. Urban hubs like Colombo and Kandy appear more lethal because they act as ``final-mile'' destinations, effectively exporting mortality from rural zones. Thus, urban $L_r$ reflects catchment burden, while low rural $L_r$ masks systemic exclusion.

In Phase 1, the 72\% unexplained variance in $G_d$ suggests that facility quantity does not guarantee accessibility. In Matale and Kalutara, road-grade constraints inflate travel times, validating terrain-blind planning is insufficient. This spatial friction drives extreme NGI variance; for instance, Mannar’s NGI (290,303.82) compared to Colombo (170.23) stems from a ``double burden'' of high $G_d$ (84.07\%) and a low resource score (17.1).

\item \textbf{Disease-Specific NGI: Infrastructure Activation and Specialist Scarcity}\\
Extreme disease score gaps, such as Moneragala’s Trauma value ($52.0 \times 10^6$), reveal that infrastructure lacks functional capacity without a specialist workforce. Since the resource score (0-1 scale) prioritizes expertise, the absence of neurosurgeons, cardiologists, or chest physicians drives massive systemic pressure. In Puttalam (Heart Disease: $6.6 \times 10^6$) and Kurunegala (Asthma: $3.6 \times 10^6$), beds remain underutilized as life-saving procedures like catheterization require specialists. In these contexts, hospitals become “institutional mirages”, physically present and accessible, yet clinically unable to provide definitive treatment.
\end{enumerate}

\section{Limitations}
Several limitations should be noted. Emergency case volumes include non-emergency presentations, so future work should restrict datasets to acute admissions for greater precision. It should also be noted that the epidemiological dataset includes non-emergency admissions, which may artificially inflate NGI values. Filtering elective and non-acute cases from the Annual Health Bulletin data would improve baseline demand accuracy in future iterations. The NGI is intentionally sensitive to specialized capital; thus, the $10^7$-scale spikes observed in districts such as Moneragala reflect a critical scarcity signal of systemic human-capital collapse rather than numerical instability. The model also relies on static OSRM-derived travel times without accounting for real-time traffic or monsoon-related road disruptions, particularly in dense urban clusters like Colombo where peak congestion alters the 60-minute travel window. Finally, $L_r$ is limited to hospital-recorded events and may underrepresent community-level burden in highly isolated, high-$G_d$ regions. Furthermore, because $L_r$ only counts hospital-recorded fatalities, it fails to capture individuals who pass away before reaching a facility. This data limitation likely underestimates the true mortality rate in underserved rural areas.

\section{Conclusion And Future Work}

This study suggests that Sri Lanka’s emergency resilience depends on aligning geographic access with appropriate medical expertise. By validating hypotheses H1-H4, the findings indicate that specialists are the primary drivers of healthcare system effectiveness; in their absence, hospitals may remain structurally present but functionally limited. The study advocates equitable decentralization, proposing that placing specialists within the most excluded 28\% of regions (``Red Clusters'') would enable patients to receive timely treatment and stabilization closer to home. This, in turn, could relieve systemic pressure by reducing patient overflow into central hubs such as Colombo and Anuradhapura, thereby easing the burden on urban referral hospitals.

Ultimately, addressing these regional gaps increases national efficiency by 9.65\%. Future extensions will integrate Google Maps Traffic API data to build a time-dependent speed model, allowing routing algorithms to dynamically capture hourly urban congestion patterns. This framework will combine with Dijkstra’s algorithm to guide patients to the best hospital in real time. Additionally, tracking brain drain and connecting this data with ``Suwa Seriya'' logistics will help build an early-warning system to protect the network, ensuring Sri Lankan healthcare functions as a synchronized system for survival.

{\footnotesize
\bibliographystyle{IEEEtranN}
\bibliography{references}
}

\end{document}